\documentclass{article}

\usepackage{microtype}
\usepackage{graphicx}
\usepackage{subcaption}
\usepackage{booktabs} % for professional tables

\usepackage{hyperref}

\usepackage[preprint]{icml2026}

\usepackage{amsmath}
\usepackage{amssymb}
\usepackage{mathtools}
\usepackage{amsthm}
\usepackage{amsfonts, bm}
\usepackage{multirow}

\usepackage[capitalize,noabbrev]{cleveref}

\theoremstyle{plain}
\newtheorem{theorem}{Theorem}[section]
\newtheorem{proposition}[theorem]{Proposition}
\newtheorem{lemma}[theorem]{Lemma}

\theoremstyle{definition}
\newtheorem{definition}[theorem]{Definition}

\theoremstyle{remark}

\usepackage[textsize=tiny]{todonotes}

\icmltitlerunning{LoFT: Long-Tailed Semi-Supervised Learning via Parameter-Efficient Fine-Tuning in Open-World Scenarios}

\begin{document}

\twocolumn[
  \icmltitle{LoFT: Long-Tailed Semi-Supervised Learning via Parameter-Efficient Fine-Tuning in Open-World Scenarios}



  \icmlsetsymbol{equal}{*}

  \begin{icmlauthorlist}
    \icmlauthor{Zhiyuan Huang}{equal,yyy}
    \icmlauthor{Jiahao Chen}{equal,yyy}
    \icmlauthor{Bing Su}{yyy}
  \end{icmlauthorlist}

  \icmlaffiliation{yyy}{Renmin University of China}

  \icmlcorrespondingauthor{Zhiyuan Huang}{huangzhiyuan@ruc.edu.cn}
  \icmlcorrespondingauthor{Jiahao Chen}{nicelemon666@gmail.com}

  \icmlkeywords{Machine Learning, ICML}

  \vskip 0.3in
]



\printAffiliationsAndNotice{}  % no special notice (required even if empty)

\begin{abstract}
    Long-tailed semi-supervised learning (LTSSL) presents a formidable challenge where models must overcome the scarcity of tail samples while mitigating the noise from unreliable pseudo-labels. Most prior LTSSL methods are designed to train models from scratch, which often leads to issues such as overconfidence and low-quality pseudo-labels. To address this problem, we first theoretically prove that utilizing a foundation model significantly reduces the hypothesis complexity, which tightens the generalization bound and in turn minimizes the Balanced Posterior Error (BPE). Furthermore, we demonstrate that the feature compactness of foundation models strictly compresses the acceptance region for outliers, providing a geometric guarantee for robustness.
    Motivated by these theoretical insights, we extend LTSSL into the foundation model fine-tuning paradigm and propose a novel framework: LoFT (Long-tailed semi-supervised learning via parameter-efficient Fine-Tuning). Furthermore, we explore a more practical setting by investigating semi-supervised learning under open-world conditions, where the unlabeled data may include out-of-distribution (OOD) samples. To handle this problem, we propose LoFT-OW (LoFT under Open-World scenarios) to improve the discriminative ability. Experimental results on multiple benchmarks demonstrate that our method achieves superior performance. Code is available: \url{https://github.com/games-liker/LoFT}
\end{abstract}

\vskip -0.1in

\begin{figure}[t]
    \centering
    \includegraphics[width=1\linewidth]{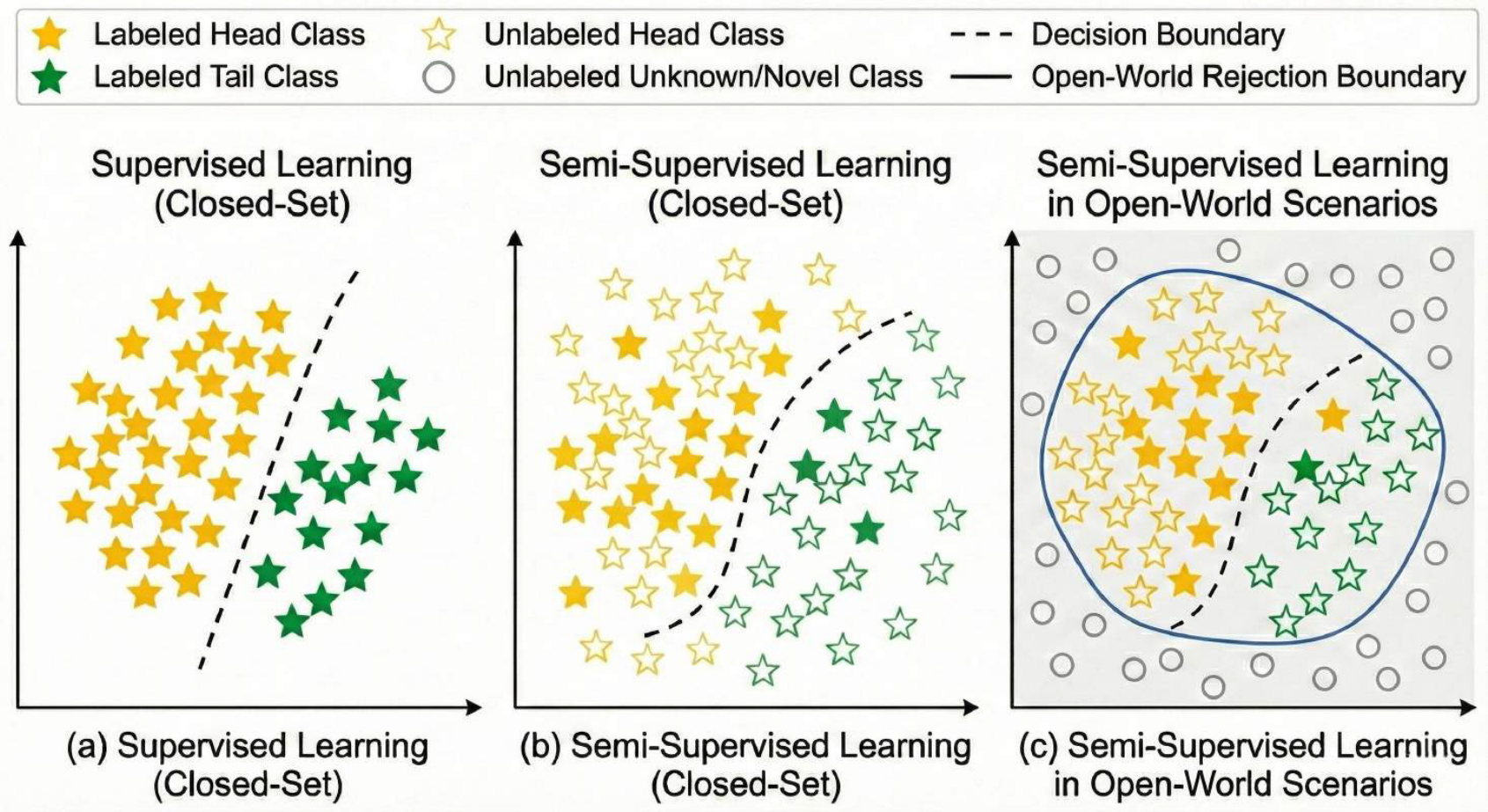}
    % \vskip -0.1in
    \caption{Differences among supervised learning, semi-supervised learning, and semi-supervised learning in open-world scenarios.}
    \label{diff}
    \vskip -0.1in
\end{figure}

\begin{figure*}[t]
  \centering
    \begin{subfigure}[t]{0.49\textwidth}\includegraphics[width=\textwidth]{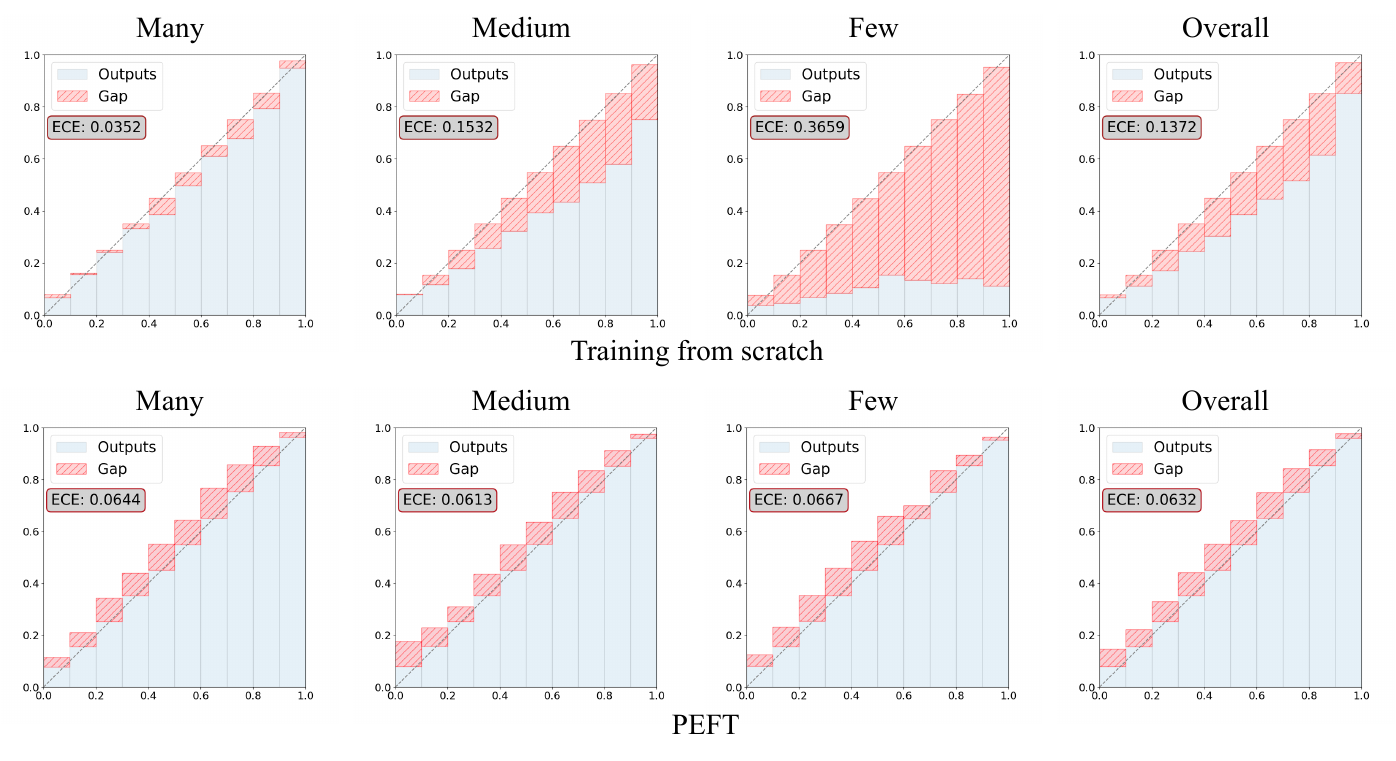}\caption{ImageNet-LT}\end{subfigure}
    \begin{subfigure}[t]{0.49\textwidth}\includegraphics[width=\textwidth]{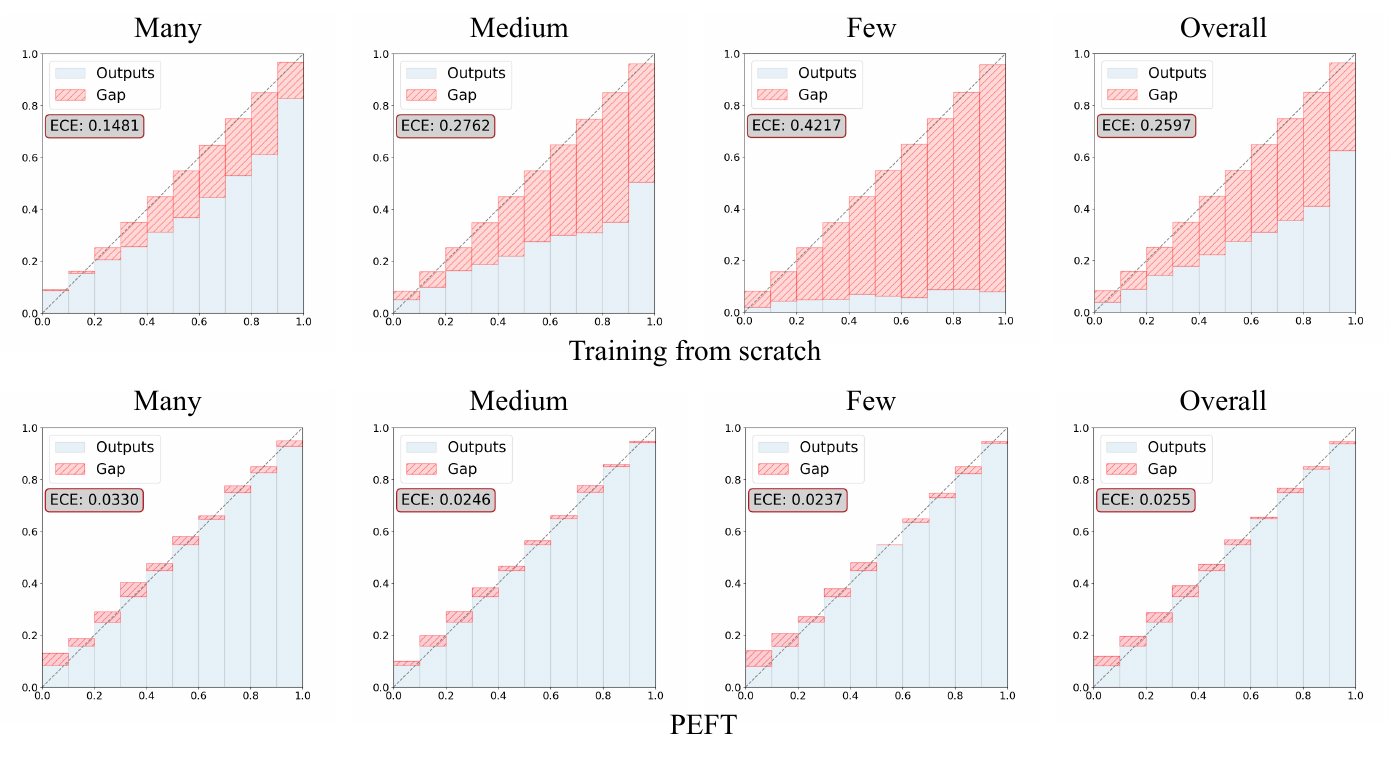}\caption{Places365-LT}\end{subfigure}
    % \vskip -0.05in
    % \vskip -0.1in
   \caption{ The reliability diagrams on (a) ImageNet-LT and (b) Places365-LT based on training from scratch and PEFT, respectively. The horizontal axis represents confidence, and the vertical axis represents accuracy. The Expected Calibration Error (ECE) is computed to quantify the calibration quality, with lower values indicating better confidence-accuracy alignment. 
   } 
   \label{merge}
   % \vskip -0.1in
\end{figure*}

\section{Introduction}

Real-world data often follows a long-tailed or imbalanced distribution, where a small number of head classes dominate the majority of samples, while the remaining tail classes are represented by only a limited number of instances~\citep{cui2019class}.
This imbalance poses significant challenges for model training, particularly in achieving satisfactory performance on tail classes.
To address this issue, LTSSL has emerged as an effective solution by incorporating a large amount of unlabeled data into the imbalanced labeled dataset~\citep{wei2021crest, Wei_2023_CVPR}.
The basic idea of LTSSL is to generate pseudo-labels for unlabeled data and select high-confidence samples to guide model training~\citep{ouali2020overview}.
While the current methods have achieved notable success and demonstrated promising results,  they still face dilemmas hindering further improvement.

Previous LTSSL approaches typically rely on training Convolutional Neural Networks (CNNs) from scratch~\citep{wei2021crest}, which presents several challenges. First, CNNs are known to be overconfident~\citep{guo2017calibration}, often assigning high-confidence scores to incorrect predictions. Although methods like FixMatch~\citep{sohn2020fixmatch} employ a ``weak-to-strong” pipeline, using weakly augmented samples to determine labels and strong augmented samples to determine the logits, this overconfidence issue persists, especially for tail classes, as shown in Fig.~\ref{merge}. Second, in the early training stages, the model produces unreliable predictions, resulting in low-quality pseudo-labels. As a result, current LTSSL approaches often require more training iterations and carefully designed strategies to dynamically manage the use of unlabeled data~\citep{Wei_2023_CVPR}. Both of the dilemmas limit the application of LTSSL.

% To address these challenges, we pivot to a theoretical perspective. We derive a generalization bound demonstrating that the massive scale of pre-training data significantly tightens the bound, effectively shielding the model from the scarcity of tail samples. Furthermore, we show that this tighter bound naturally leads to a minimized Balanced Prediction Error (BPE). Driven by these theoretical insights, we introduce a streamlined framework that leverages the high generalization capabilities of Foundation Models to achieve inherently well-calibrated and high-quality pseudo-labels.

Beyond the standard LTSSL setting~\citep{yang2020rethinking}, a critical gap exists between theoretical assumptions and reality: most methods assume distributional homogeneity, where the labeled and unlabeled sets share the same class space. To bridge this gap, we explore a more pragmatic and challenging setting: Open-World LTSSL.
In real-world scenarios, such as wildlife classification, the unlabeled stream inevitably contains Out-of-Distribution (OOD) samples (e.g., rare or unknown species) not present in the labeled set. Directly applying existing LTSSL methods in this context carries the risk of forcing OOD samples into in-distribution (ID) classes, thereby corrupting the feature space.
Moreover, models trained from scratch typically lack the semantic capacity to effectively identify or reject these OOD samples~\citep{hendrycks2016baseline}.

% To tackle this, we theoretically analyze the geometry of the feature space. We demonstrate that models trained from scratch typically suffer from feature scattering, where loose decision boundaries inadvertently encompass OOD samples. In contrast, we prove that Foundation Models exhibit inherent feature compactness, which significantly compresses the acceptance region for OOD data on the hypersphere. Motivated by this geometric insight, we propose LoFT-OW (LoFT under Open-World scenarios). By leveraging the compact feature clusters of the fine-tuned foundation model, LoFT-OW incorporates a built-in OOD detection mechanism to natively filter out irrelevant samples. This effectively mitigates negative transfer and preserves the purity of representation learning, as evidenced by the superior robustness shown in Tab.~\ref{ood_table}.

Motivated by theoretical insights and empirical observations, we introduce a streamlined framework called LoFT(Long-Tailed Semi-Supervised Learning via Parameter-Efficient Fine-Tuning) that leverages the high generalization capabilities of Foundation Models(FMs) to achieve inherently well-calibrated and high-quality pseudo-labels. Moreover, we propose LoFT-OW (LoFT under Open-World scenarios). By leveraging the compact feature clusters of the fine-tuned foundation model, LoFT-OW incorporates a built-in OOD detection mechanism to natively filter out irrelevant samples. This effectively mitigates negative transfer and preserves the purity of representation learning. %, as evidenced by the superior robustness shown in Tab.~\ref{ood_table}.

Our contributions are summarized as follows:
\begin{itemize}
    % \item Grounded in theoretical analysis, We address the LTSSL problem and propose LoFT, a novel framework that leverages Parameter-Efficient Fine-Tuning (PEFT) foundation models. Through comprehensive experiments, we analyze the confidence behavior of LoFT and observe that it is inherently well-calibrated. We further demonstrate that this property can be effectively utilized to improve the quality of pseudo-labels.
    \item We made theoretical contributions, and motivated by theoretical analysis and sufficient empirical experiments, we address the LTSSL problem and propose LoFT that is inherently well-calibrated and can be effectively utilized to impove the quality of pseudo-labels.

    \item We extend LTSSL to a more realistic Open-World Scenario, named LoFT-OW, where unlabeled data may contain OOD samples. LoFT incorporates a built-in OOD detection mechanism, filtering out irrelevant samples and improving model robustness and representation learning in diverse real-world data conditions.

    \item We conduct experiments on traditional LTSSL benchmarks, including CIFAR-LT and ImageNet127, and observe that LoFT achieves competitive performance. Furthermore, LoFT achieves superior performance in the more challenging open-world scenarios, outperforming previous methods even when using only 10\% of the unlabeled data compared with previous works, highlighting its strong discriminative capability.

\end{itemize}

\section{Related Work}

\paragraph{Long-tailed semi-supervised learning}
LTSSL~\citep{peng2023dynamic, hou2025square, wei2021crest} aims to improve the performance of models trained on long-tailed labeled data by leveraging additional unlabeled data. The basic idea is to generate pseudolabels for the unlabeled samples and incorporate them into the training process. CReST~\citep{wei2021crest} observes that models trained under imbalanced distributions can still generate high-precision pseudolabels for tail classes. Based on this insight, it proposes the class-rebalancing self-training framework to improve performance. In~\citep{Wei_2023_CVPR}, the authors relax the assumption of consistent class distributions between labeled and unlabeled data and introduce ACR, a method that dynamically refines pseudo-labels by estimating the true class distribution of unlabeled data under a unified formulation. ADELLO~\citep{sanchez2024ADELLO_LTSSL} presents FlexDA, a dynamic logit adjustment and distillation-based framework that enhances calibration and achieves strong performance in LTSSL settings. Recently, FMs~\citep{radford2021learning}, pre-trained on large-scale datasets, have demonstrated strong generalization capabilities across a variety of downstream tasks, including those with long-tailed distributions~\citep{shi2024long, tian2022vl, dong2022lpt}. However, how to effectively leverage FMs to benefit LTSSL remains an open and underexplored research direction.  In this paper, we aim to address this challenge and propose LoFT, a novel framework designed to integrate the strengths of FMs into the LTSSL paradigm.

\paragraph{Long-tailed Confidence calibration}
Confidence calibration aims to align the predicted confidence scores with the true accuracy, which is important for safety measurement, OOD detection~\citep{liu2024rethinking}. Prior studies have shown that modern CNNs tend to be overconfident~\citep{Tomani_2021_CVPR, guo2017calibration}, particularly under long-tailed distributions~\citep{zhong2021improving}. MiSLAS~\citep{zhong2021improving} addresses this issue by introducing a two-stage training pipeline that incorporates three key techniques: mixup~\citep{zhang2017mixup} pre-training, label-aware smoothing, and batch normalization~\citep{ioffe2015batch} shifting. These techniques collectively enhance the model's calibration capability. UniMix~\citep{xu2021towards} extends the mixup strategy to imbalanced scenarios by adopting an advanced mixing factor and a sampling strategy that favors minority classes, thereby improving calibration performance under long-tailed distributions. Recently, adapting FMs to imbalanced learning has attracted increasing attention. However, the issue of confidence calibration in this setting remains largely underexplored. As previously discussed, a well-calibrated model is crucial for generating high quality pseudo-labels, which are essential for effective semi-supervised learning. In this work, we investigate confidence calibration within the context of LTSSL to further enhance performance under long-tailed distributions.

\begin{table}[t]
\centering

\setlength{\tabcolsep}{4pt}
\small
\caption{The results on OOD tasks on different datasets. PEFT$^\dag$ and PEFT$^\ddag$ denote the fine-tuned model from CLIP and OpenCLIP, respectively.}
% \vskip -0.1in
\label{ood_table}
\tabcolsep=0.16cm
\begin{tabular}{c|c|c|c|c|c}
\toprule
% \multicolumn{6}{|c|}{CIFAR100-LT} & 
% \hline
OOD Dataset & 
Method & 
AUC$\uparrow$ & AP-in$\uparrow$ & AP-out$\uparrow$ & FPR$\downarrow$ \\
\midrule

\multirow{4}{*}{Texture}
  & OE & 76.01 & 85.28 & 57.47 & 87.45 \\
  & OCL & 75.92 & 82.99 & 66.48 & 70.01 \\
\cmidrule{2-6}
  & PEFT$^\dag$ & \underline{87.86} & \underline{92.79} & \underline{80.15} & \underline{49.45} \\
  & PEFT$^\ddag$ & \textbf{91.32} & \textbf{94.66} & \textbf{86.22} & \textbf{38.26} \\
\midrule

\multirow{4}{*}{SVHN}
  & OE & 81.82 & 73.25 & 89.10 & 80.98 \\
  & OCL & 78.64 & 69.21 & 86.26 & 86.38 \\
\cmidrule{2-6}
  &  PEFT$^\dag$ & \underline{86.62} & \underline{73.87} & \underline{94.26} & \underline{47.29} \\
  & PEFT$^\ddag$ &\textbf{90.68} & \textbf{81.80} & \textbf{95.98} & \textbf{41.00} \\
\midrule

\multirow{4}{*}{CIFAR-10}
  & OE & 62.60 & 66.16 & 57.77 & 93.53 \\
  & OCL & 60.29 & 63.21 & 55.71 & 94.22 \\
\cmidrule{2-6}
  &  PEFT$^\dag$ & \underline{83.97} & \underline{84.42} & \underline{82.61} & \underline{61.98} \\
  & PEFT$^\ddag$ & \textbf{86.39} & \textbf{86.95} & \textbf{85.38} & \textbf{57.38} \\
\midrule

\multirow{4}{*}{TinyImageNet}
  & OE & 68.22 & 79.36 & 51.82 & 88.54 \\
  & OCL & 69.56 & 79.97 & 54.47 & 85.91 \\
\cmidrule{2-6}
  &  PEFT$^\dag$ & \underline{81.34} & \underline{88.30} & \underline{70.20} & \underline{70.03} \\
  & PEFT$^\ddag$ & \textbf{83.35} & \textbf{89.85} & \textbf{72.98} & \textbf{66.02} \\
\midrule

\multirow{4}{*}{LSUN}
  & OE & 76.81 & 85.33 & 60.94 & 83.79 \\
  & OCL & \underline{79.14} & \underline{86.56} & \underline{66.58} & \underline{75.07} \\
\cmidrule{2-6}
  &  PEFT$^\dag$ & 78.16 & 86.32 & 65.86 & 75.45 \\
  & PEFT$^\ddag$ & \textbf{81.29} & \textbf{88.45} & \textbf{70.49} & \textbf{69.50} \\
\midrule

\multirow{4}{*}{Place365}
  & OE & 75.68 & 60.99 & 86.51 & 83.55 \\
  & OCL & 77.81 & 62.80 & 88.39 & 79.97 \\
\cmidrule{2-6}
  &  PEFT$^\dag$ & \underline{84.65} & \underline{71.67} & \underline{93.00} & \underline{58.36} \\
  & PEFT$^\ddag$ & \textbf{86.04} & \textbf{74.25} & \textbf{93.65} & \textbf{55.43} \\
\midrule

\multirow{4}{*}{Average}
  & OE & 73.52 & 75.06 & 67.27 & 86.30 \\
  & OCL & 73.56 & 74.12 & 69.65 & 81.93 \\
\cmidrule{2-6}
  &  PEFT$^\dag$ & \underline{83.77} & \underline{82.90} & \underline{81.01} & \underline{60.43} \\
  & PEFT$^\ddag$ & \textbf{86.51} & \textbf{85.99} & \textbf{84.12} & \textbf{54.60} \\
\bottomrule

\end{tabular}

\end{table}

\begin{figure*}[t]
  \centering
{\includegraphics[width=1.0\textwidth]{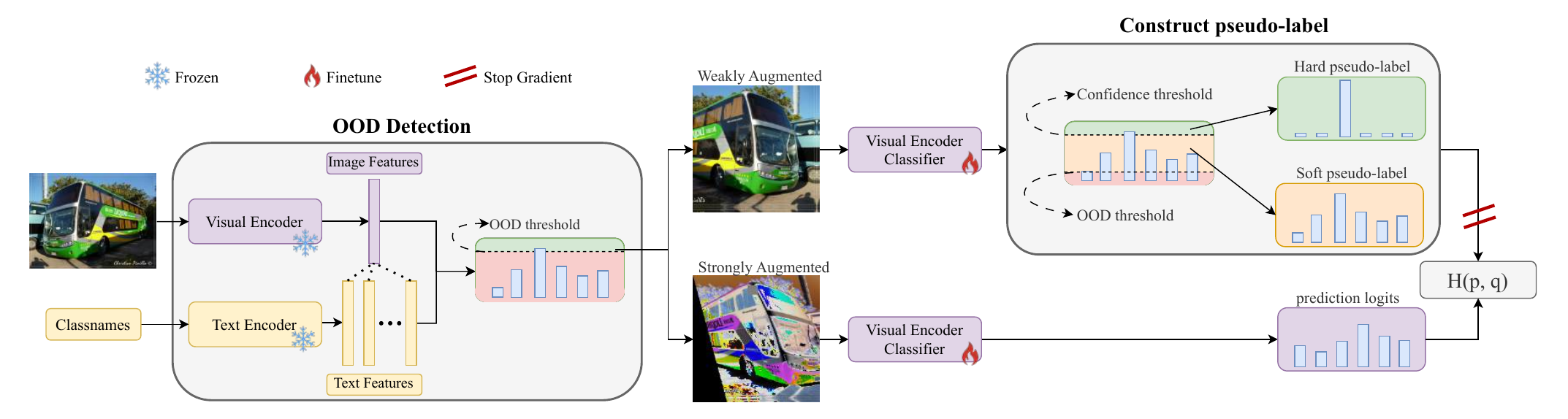}}
   \caption{ Illustration of the proposed LoFT-OW. $H(p, q)$ denotes the cross-entropy.
   } 
   \label{main}
   % \vskip -0.1in
\end{figure*}

\section{Theoretical Motivation and Observations}
\label{sec:theory}

% Current LTSSL methods predominantly rely on training deep models from scratch. While effective to a degree, this paradigm suffers from a ``vicious cycle'': the scarcity of tail samples leads to high generalization error, which induces severe overconfidence and high Balanced Posterior Error (BPE), ultimately resulting in unreliable pseudo-labels that mislead the self-training process.

% % In this section, we provide a unified theoretical framework to explain why fine-tuning Foundation Models (FMs) via LoFT fundamentally breaks this cycle. We formalize the generalization advantage of pre-training through the lens of \textit{hypothesis complexity reduction} and prove how it propagates to minimize the target risk on unlabeled data.

% In this section, we first present a theoretical analysis, showing that FMs can address the aforementioned problem. Furthermore, we validate the theoretical analysis through comprehensive experiments, which also inspire the design of LoFT and LoFT-OW.

Current LTSSL methods predominantly rely on training deep models from scratch. While effective to a degree, this paradigm suffers from a ``vicious cycle'': the scarcity of tail samples leads to high generalization error, which induces severe overconfidence and high Balanced Posterior Error (BPE), ultimately resulting in unreliable pseudo-labels that mislead the self-training process.

In this section, we provide a unified theoretical framework to explain how fine-tuning FMs via PEFT fundamentally breaks this cycle. We structure our analysis into three logical steps: 
% \vskip -0.1in
\begin{enumerate} 
    \vskip -0.1in
    \item We first introduce a generalization bound based on hypothesis complexity (Lemma~\ref{lemma:gen_bound}), proving that limiting the search space via PEFT can mathematically compensate for the scarcity of tail samples. 
    \vskip -0.1in
    \item We then bridge this bound to the BPE (Theorem~\ref{thm:loft_main}), demonstrating that better generalization directly translates to lower worst-case risk and more reliable pseudo-labels. 
    \vskip -0.1in
    \item Finally, we extend the analysis to Open-World scenarios (Proposition~\ref{prop:ood}), showing that the geometric compactness of pre-trained features inherently facilitates the rejection of OOD noise. 
    \vskip -0.1in
\end{enumerate} 
% \vskip -0.1in
These theoretical insights are further validated by empirical observations, which collectively inspire the design of LoFT and LoFT-OW. The detailed proofs are in the Appendix.

\subsection{Theoretical Analysis}

% Let $\mathcal{D}_S$ and $\mathcal{D}_U$ denote the distributions of the labeled source set (with size $N_S$) and the unlabeled target set, respectively. Let $h_{scr}$ denote a model trained from scratch, and $h_{peft}$ denote a model initialized from pre-training and fine-tuned via PEFT.

Here, we formally compare the learning dynamics of a model trained from scratch, denoted as $h_{scr}$, versus a model initialized from pre-training and fine-tuned via PEFT, denoted as $h_{peft}$. Let $\mathcal{D}_S$ (with size $N_S$) and $\mathcal{D}_U$ denote the distributions of the labeled source set and the unlabeled target set, respectively.
Our analysis proceeds by examining how the Hypothesis Complexity determines the upper bound of generalization error, and how this bound dictates the reliability of the model under long-tailed and open-world settings.

\paragraph{Generalization Error and Hypothesis Complexity}

Standard generalization bounds depend on both the sample size and the complexity of the hypothesis class. For long-tailed data, the sample size $N_S^{(y)}$ for a tail class $y$ is negligible, causing the error bound for scratch models (which have high complexity) to become vacuously loose. The following lemma illustrates that PEFT tightens this bound by constraining the hypothesis space, effectively compensating for the lack of data with prior knowledge.

% \vspace{-0.1in}

\begin{lemma}[Generalization Bound via Complexity Reduction]
\label{lemma:gen_bound}
Let $\mathfrak{R}(\mathcal{H})$ denote the Rademacher complexity of the hypothesis class. The generalization error $\mathcal{R}_S(h)$ is bounded with probability at least $1-\delta$.

For \textbf{Training from Scratch}, the model searches a vast hypothesis space $\mathcal{H}_{scr}$ (e.g., all possible CNN weights). The bound for a specific class $y$ is dominated by the ratio of high complexity to scarce samples:
\begin{equation}
    \mathcal{R}_S(h_{scr}|y) \le \hat{\mathcal{R}}_S(h_{scr}|y) + \mathcal{O}\left(\frac{\mathfrak{R}(\mathcal{H}_{scr})}{\sqrt{N_S^{(y)}}}\right).
\end{equation}
For tail classes where $N_S^{(y)} \to 0$, the uncertainty term explodes.

In contrast, PEFT constrains the search to a significantly smaller subspace $\mathcal{H}_{peft}$ (e.g., only head or adapter parameters), conditioned on robust pre-trained features. This drastically reduces the complexity term $\mathfrak{R}(\mathcal{H}_{peft}) \ll \mathfrak{R}(\mathcal{H}_{scr})$.
\begin{equation}
    \mathcal{R}_S(h_{peft}|y) \le \hat{\mathcal{R}}_{S}(h_{peft}|y) + \epsilon_{trans} + \mathcal{O}\left(\frac{\mathfrak{R}(\mathcal{H}_{peft})}{\sqrt{N_S^{(y)}}}\right),
\end{equation}
where $\epsilon_{trans}$ is the transfer error (assumed small for FMs). Even with small $N_S^{(y)}$, the significantly reduced numerator ensures a tight bound, effectively compensating for sample scarcity.
\end{lemma}

\paragraph{Bridging Generalization to BPE}

% The core challenge in LTSSL is minimizing the \textit{Balanced Posterior Error} (BPE)~\cite{wei2024learning}, which measures the worst-case error across classes: $\text{BPE}(h) = \max_y \mathbb{P}(h(x) \neq y | Y=y)$.
Having established that PEFT yields a tighter generalization bound, we now connect this result to the core challenge of LTSSL: the \textit{BPE}~\cite{wei2024learning}. The following theorem explains that by minimizing the worst-case generalization error across classes, PEFT specifically prevents the ``collapse'' of performance on tail classes.

\begin{theorem}[Superiority of PEFT on BPE]
\label{thm:loft_main}
Assuming the transfer error $\epsilon_{trans}$ is negligible compared to the complexity gap, the PEFT guarantees a lower upper bound on the BPE compared to training from scratch.

\textit{Proof Sketch.} The BPE is explicitly determined by the worst-case class-conditional risk: $\text{BPE}(h) = \max_y \mathcal{R}_S(h|y)$.
By Lemma~\ref{lemma:gen_bound}, for tail classes, the risk upper bound for $h_{scr}$ is loose due to the high $\mathfrak{R}(\mathcal{H}_{scr})$. In contrast, $h_{peft}$ enforces a tight bound for all classes $y$ by minimizing the hypothesis complexity. Consequently, the maximum risk over all classes is strictly reduced:
\begin{equation}
\begin{aligned}
    \max_y \mathcal{R}_S(h_{peft}|y) \ll \max_y \mathcal{R}_S(h_{scr}|y) \\
    \implies \text{BPE}(h_{peft}) \ll \text{BPE}(h_{scr}).
\end{aligned}
\end{equation}
This lower BPE ensures more reliable pseudo-labels for tail classes, establishing a virtuous cycle for self-training.
\end{theorem}

\paragraph{Theoretical Insight for Open-World Scenarios}
\label{sec:theory_ood}

% A key challenge in Open-World LTSSL is distinguishing OOD samples from tail classes. We justify the robustness of PEFT via the geometry of the feature space on the hypersphere.

Finally, beyond the closed-set error, we analyze the capability to distinguish OOD samples. We justify the robustness of PEFT via the geometry of the feature space, showing that a compact feature distribution is mathematically equivalent to a stronger rejection capability against open-world noise.

\begin{proposition}[OOD Robustness via Feature Concentration]
\label{prop:ood}
Let the feature space be the unit hypersphere $\mathbb{S}^{d-1}$. For a normalized linear classifier (equivalent to a nearest-centroid classifier), the acceptance region for class $k$ is a spherical cap defined by an angular threshold $\theta_k$.

Models trained from scratch typically suffer from \textit{feature scattering}, resulting in loose decision boundaries (large $\theta_k^{scr}$). The probability of a random OOD sample $x_{out}$ falling into this region is proportional to the volume of the spherical cap. By the concentration of measure on the sphere~\citep{vershynin2018high}:
\begin{equation}
    \mathbb{P}(f(x_{out}) \in \text{Cap}(\theta_k)) \le \exp\left(-\frac{d}{2}\cos^2\theta_k\right).
\end{equation}
FMs trained with contrastive objectives (like CLIP) exhibit \textbf{Feature Compactness}~\citep{wang2020understanding}, enforcing tight intra-class clustering (small $\theta_k^{loft}$). Since the volume decays exponentially with $\cos^2\theta$, a small reduction in angular spread leads to a massive reduction in OOD acceptance probability:
\begin{equation}
    \mathbb{P}(f(x_{out}) \in \text{Cap}(\theta_k^{peft})) \ll \mathbb{P}(f(x_{out}) \in \text{Cap}(\theta_k^{scr})).
\end{equation}
This geometric property allows FMs to effectively filter OOD noise using confidence thresholding.
\end{proposition}

\subsection{Empirical Observations}

% We validate our theoretical motivation with empirical studies on confidence calibration and open-world robustness.
To substantiate the theoretical claims, we provide empirical evidence focusing on two key aspects: confidence calibration (corroborating the reduced BPE in Theorem~\ref{thm:loft_main}) and open-world robustness (verifying the geometric compactness in Proposition~\ref{prop:ood}).

\paragraph{Confidence Calibration.}
Calibration serves as a proxy for validating the BPE reduction.
According to Theorem~\ref{thm:loft_main}, pre-training should suppress the conditional risk on tail classes by reducing hypothesis complexity.
As shown in Fig.~\ref{merge}, we visualize the confidence--accuracy diagram on ImageNet-LT and Places365-LT. Following previous works~\citep{liu2019large}, we divide the classes into three groups, ``Many'', ``Medium'', and ``Few'', based on the number of training samples per class. We observe that models trained from scratch tend to exhibit significant overconfidence on the unseen test set, particularly for the tail classes. Specifically, the scratch-trained model yields an ECE of $0.1372$ across the entire dataset. Moreover, the tail classes suffer from more pronounced overconfidence compared to head classes. In contrast, models fine-tuned using PEFT demonstrate substantially improved calibration, with tail classes no longer exhibiting such severe overconfidence. We attribute this improvement to the extensive pretraining of FMs on large-scale data, which reduces model uncertainty and enhances calibration. Additionally, PEFT modifies only a small subset of parameters, thereby preserving the generalization capabilities of the FMs while effectively adapting to the target task.

\paragraph{Open-World Robustness.}

As shown in Tab.~\ref{ood_table}, we fine-tune the FMs on CIFAR-100-LT and evaluate its performance on a variety of OOD datasets, including SVHN~\citep{goodfellow2013multi}, CIFAR-10~\citep{krizhevsky2009learning}, Tiny ImageNet~\citep{le2015tiny}, LSUN~\citep{yu2015lsun}, and Places365~\citep{zhou2017places}. We adopt the Maximum Softmax Probability (MSP)~\citep{hendrycks2016baseline} as the OOD detection strategy and compare our approach against baseline methods, including OE~\citep{hendrycks2018deep} and OCL~\citep{miao2024out}. Across multiple evaluation metrics, the model fine-tuned on OpenCLIP achieves the best overall performance, with an average score of $86.51$ across the six datasets.
These results validate Proposition~\ref{prop:ood}: the compact feature space of FMs create tighter spherical acceptance regions, natively filtering out OOD noise.

\section{Method}

% Movitaed by theoretical analysis and empirical observations, we propse LoFT and LoFT-OW to solve LTSSL and Open-World scenarios problems that previous methods struggle.

Building on the analysis in Sec.~\ref{sec:theory}, we propose \textbf{LoFT} and its open-world extension, \textbf{LoFT-OW}. The theoretical guarantee of reduced BPE motivates LoFT to leverage the superior calibration of FMs to employ a confidence-aware self-training strategy, leveraging improved calibration to assign reliable hard and soft pseudo-labels. Meanwhile, the geometric compactness of pre-trained features inspires LoFT-OW, which utilizes a dual-stage filtering mechanism to effectively reject OOD samples. The details are formulated below.

\subsection{LoFT}
In modern LTSSL, models are typically optimized by jointly minimizing a supervised classification loss on labeled data, used to learn initial discriminative representations, and a regularization loss on unlabeled data, which further refines the learned features and enhances generalization.

For the supervised classification loss, we adopt the Logit Adjustment~\citep{menon2020long} as the criterion on the labeled long-tailed dataset. The optimization objective is:
\begin{equation}
    \mathcal{L}_s 
    = \frac{1}{\mid \mathcal{D}_S \mid} \sum_{\bm{x} \sim \mathcal{D}_S} 
    H\Bigl(y_b,\;f\bigl(\mathcal{W}(\bm{x})\bigr) + \tau\,\log \mathbb{P}_{S}(Y)\Bigr),
\end{equation}
where \(\mathcal{W}(\cdot)\) denotes a weak augmentation operation (e.g., random crop or horizontal flip), \(\tau\) is a scaling hyperparameter, and \(\mathbb{P}_{S}(Y)\) represents the empirical class prior estimated from the labeled dataset.

For the regularization loss on unlabeled samples, we follow the basic principle from prior work~\citep{sohn2020fixmatch}, where a weakly augmented view is used to generate pseudo-labels, and a strongly augmented view is used to obtain logits for optimization. To better handle uncertain predictions, we partition unlabeled samples into high-confidence and low-confidence subsets based on their MSP, and apply different optimization strategies accordingly. Specifically, we define a binary mask $M_{\bm{x}}$ to indicate whether an unlabeled sample is considered high-confidence, computed as:
\begin{equation}
    M_{\bm{x}} =
    \begin{cases}
        1, & \quad \mathrm{MSP}(\bm{x}) > c_u \\
        0, & \quad \mathrm{MSP}(\bm{x}) \leq c_u \\
    \end{cases}
    \label{m_x}
\end{equation}

The optimization objective for unlabeled samples is:
\begin{equation}
\begin{aligned}
    \mathcal{L}_u 
    = \frac{1}{\mid \mathcal{D}_U \mid} 
    &\sum_{\bm{x} \sim \mathcal{D}_U } \lambda_1 M_{\bm{x}} \cdot H\bigl(\hat y,\,f(\mathcal{A}(\bm{x}))\bigr) \\
    &+ \lambda_2 (1 - M_{\bm{x}}) \cdot H\bigl(f(\mathcal{W}(\bm{x})),\,f(\mathcal{A}(\bm{x}))\bigr),
\end{aligned}
\label{unlabel_ot}
\end{equation}
where \(\hat y = \arg\max f(\mathcal{W}(\bm{x}))\) denotes the hard pseudo-label derived from the weakly augmented view, and $\mathcal{A}(\cdot)$ denotes a strong augmentation. $\lambda_1$ and $\lambda_2$ are  hyperparameters. 

In Eq.~\ref{unlabel_ot}, for high-confidence samples ($M_{\bm{x}} = 1$), we apply hard pseudo-labels by assigning the most probable class using the model's prediction. For low-confidence samples ($M_{\bm{x}} = 0$), we apply soft pseudo-labels by leveraging the full predicted probability distribution, which provides smoother supervision and better captures prediction uncertainty. We analyze that, as shown in Fig.~\ref{merge}, under our fine-tuning framework, the model's confidence score is strongly correlated with prediction accuracy. Since high-confidence samples are generally more reliable, we apply hard supervision to them, while soft supervision is used for low-confidence samples to mitigate overfitting and enhance generalization. Furthermore, as discussed previously, our fine-tuned model exhibits better calibration for tail classes compared to models trained from scratch. Consequently, we do not distinguish between head and tail classes when determining the confidence mask in Eq.~\ref{m_x}, e.g., setting different thresholds for head or tail classes, which also reduces the number of required hyper-parameters. Finally, the overall training objective is:

\vspace{-0.5cm}

\begin{equation}
    \mathcal{L} \;=\; \mathcal{L}_s \;+\; \,\mathcal{L}_u \;
\end{equation}

\subsection{LoFT-OW (LoFT under Open-World scenarios)}
Traditional LTSSL methods typically assume that all unlabeled data originates from the same distribution as the labeled data—a condition that rarely holds in real-world scenarios. In practice, unlabeled data are often collected from broad, unconstrained sources such as the web or dynamic field environments, where it is highly likely that a substantial portion of samples lie outside the distribution of the predefined labeled classes. These OOD samples, if not properly handled, can degrade model performance by introducing misleading supervision. To address this challenge, we propose an extension of our framework to open-world settings, termed LoFT-OW (LoFT under Open-World scenarios). LoFT-OW is designed to effectively detect and filter out OOD samples during training, thereby mitigating their adverse effects and enhancing performance in long-tailed, semi-supervised learning.

As shown in Fig.~\ref{main}, we adopt a two-stage filtering strategy to identify OOD samples. In the first stage, we employ a zero-shot filtering mechanism, where the foundation model assigns confidence scores to each unlabeled sample. Only those with confidence exceeding a high-confidence threshold \(t_{\mathrm{HC}}\) are retained, resulting in a cleaner and more reliable pseudo-labeled subset, denoted as \(\widetilde{\mathcal{D}}_U\). This filtered dataset is typically smaller in size and can be leveraged for subsequent fine-tuning. Beyond this initial stage, we further exploit the strong OOD detection capability of the fine-tuned model, which has been verified previously. We define the filtering function as follows:

\begin{equation}
    M_{\bm{x}}^{ood} =
    \begin{cases}
        1, & \quad \mathrm{MSP}(\bm{x}) > c_{ood} \\
        0, & \quad \mathrm{MSP}(\bm{x}) \leq c_{ood} \\
    \end{cases} ,
    \label{m_x2}
\end{equation}

where $c_{ood}$ is the hyper-parameter to control the filtering strength. Then the optimization object for the unlabeled set under open-world scenarios is:
\begin{equation}
\begin{aligned}
    \mathcal{L}_u 
    = & \frac{1} {\mid \widetilde{\mathcal{D}}_U \mid}  \sum_{\bm{x} \sim \widetilde{\mathcal{D}}_U }
    \lambda_1M_{\bm{x}}^{ood} M_{\bm{x}} \cdot H\bigl(\hat y,\,f(\mathcal{A}(\bm{x}))\bigr) \\
    &+ \lambda_2 M_{\bm{x}}^{ood}(1 - M_{\bm{x}}) \cdot H\bigl(f(\mathcal{W}(\bm{x})),\,f(\mathcal{A}(\bm{x}))\bigr),
\end{aligned}
\label{unlabel_ot2}
\end{equation}

\begin{table*}[t]
\centering
\small
\caption{The accuracy results on CIFAR-100-LT with different hyper-parameters of $\gamma_u$ and $\gamma_l$. PEFT refers to the fine-tuning method of LoFT using only supervised data that demonstrates the capabilities of the foundation model when using only labeled data. The comparison proves the performance improvement achieved by utilizing unlabeled data within the LoFT and LoFT-OW framework.}
\vskip -0.1in
\setlength{\tabcolsep}{4.6pt}
\begin{tabular}{l|c|c|c|c|c|c|c|c|c}
\toprule
% \multicolumn{6}{|c|}{CIFAR100-LT} & 
% \hline
% \multicolumn{2}{c}{} & \multicolumn{8}{c}{CIFAR-100-LT} \\
% \midrule

\multicolumn{2}{c}{\multirow{3}{*}{Method}}
  & \multicolumn{2}{c}{$\gamma=\gamma_l=\gamma_u=10$} & \multicolumn{2}{c}{$\gamma=\gamma_l=\gamma_u=20$} & \multicolumn{2}{c}{$\gamma_u=1$ (uniform)} & \multicolumn{2}{c}{$\gamma_u=1/10$ (reversed)} \\
  
\cmidrule{3-10}
  \multicolumn{2}{c}{} & $N_1=50$ & $N_1=150$ & $N_1=50$ & $N_1=150$ & $N_1=50$ & $N_1=150$ & $N_1=50$ & $N_1=150$ \\
  \multicolumn{2}{c}{} & $M_1=400$ & $M_1=300$ & $M_1=400$ & $M_1=300$ & $M_1=400$ & $M_1=300$ & $M_C=400$ & $M_C=300$ \\
\midrule

\multicolumn{2}{l|}{FixMatch} & 45.2 & 56.5 & 40.0 & 50.7 & 45.5 & 58.1 & 44.2 & 57.3 \\
\cmidrule{3-10}

\multicolumn{2}{l|}{+ACR} & 55.7 & 65.6 & 48.0 & 58.9 & 66.0 & 73.4 & 57.0 & 67.6 \\
\cmidrule{3-10}

\multicolumn{2}{l|}{+ACR+BEM} & 55.8 & 66.3 & 48.6 & 59.8 & - & - & - & - \\
\cmidrule{3-10}

\multicolumn{2}{l|}{+TCBC} & - & 59.4 & - & 53.9 & - & 63.2 & - & 59.9 \\
\cmidrule{3-10}

\multicolumn{2}{l|}{+CPE} & 50.3 & 59.8 & 43.8 & 55.6 & - & - & - & 60.8 \\
\cmidrule{3-10}

\multicolumn{2}{l|}{+CCL} & 53.5 & 63.5 & 46.8 & 57.5 & 59.8 & 67.9 & 54.4 & 64.7 \\
\midrule

\multirow{3}{*}{CLIP}
  & PEFT & 75.5 & 79.7 & 74.0 & 78.4 & 75.5 & 79.7 & 75.5 & 79.7  \\
  & LoFT & 78.8 & 81.1 & 75.3 & 79.3 & 78.0 & 81.0 & 77.3 & 80.6  \\
  & LoFT-OW & 76.5 & 79.9 & 73.6 & 78.6 & 76.6 & 80.0 & 76.4 & 80.0 \\
\midrule

\multirow{3}{*}{OpenCLIP}
  & PEFT & 78.0 & \underline{81.7} & 75.3 & \underline{81.1} & 78.0 & 81.7 & 78.0 & 81.7  \\
  & LoFT & \textbf{81.8} & \textbf{83.2} & \textbf{78.4} & \textbf{81.2} & \textbf{80.3} & \textbf{83.6} & \textbf{79.8} & \textbf{82.3} \\
  & LoFT-OW & \underline{79.3} & 81.6 & \underline{75.4} & 80.8 & \underline{78.6} & \underline{82.1} & \underline{79.7} & \underline{82.0} \\
\bottomrule

\end{tabular}
% }

\label{cifar_res}
\end{table*}

\begin{table}[t]
\centering
\setlength{\tabcolsep}{4.6pt}
\small
\caption{The results on ImageNet-127. PEFT refers to the fine-tuning method of LoFT using only supervised data.}
\vskip -0.1in
% \resizebox{0.5\textwidth}{!}{
\begin{tabular}{l|c|c|c}
\toprule
\multicolumn{2}{c|}{Method} & training iterations & Accuracy \\
\midrule

\multicolumn{2}{l|}{FixMatch} &250000 & 42.3 \\

\multicolumn{2}{l|}{+BEM} &250000 & 58.2 \\

\multicolumn{2}{l|}{+ACR} &250000 & 63.6 \\

\multicolumn{2}{l|}{+ACR+BEM} &250000 & 63.9 \\

\multicolumn{2}{l|}{+CCL} &250000 & 67.8 \\

\midrule

\multirow{3}{*}{CLIP}
  & PEFT &10000 & 71.7 \\
  & LoFT &10000 & 73.3 \\
  & LoFT-OW &10000 & 73.1 \\
\midrule

\multirow{3}{*}{OpenCLIP}
  & PEFT &10000 & 72.5 \\
  & LoFT &10000& \underline{73.9} \\
  & LoFT-OW &10000& \textbf{74.2} \\
\bottomrule

\end{tabular}

\label{imagenet_res}
\vskip -0.1in
\end{table}

\section{Experiments}

\subsection{Experimental Setup}
To validate the efficacy of our method under long-tailed distributions and in open-world semi-supervised learning scenarios, we conduct experiments on two long-tailed benchmarks: CIFAR-100-LT~\citep{cui2019class}, ImageNet-127~\citep{wei2021crest}. For ImageNet-127, we only use 10\% of the unlabeled data compared with ACR. For CIFAR-100-LT, let $N_k$ denote the number of labeled samples for class $k$, with $N_1 \geq N_2 \geq \cdots \geq N_K$. The imbalance ratio of the labeled dataset is defined as $\gamma_l = \frac{N_1}{N_C}$.  Similarly, let $M_c$ denote the number of unlabeled samples for class $c$, and the imbalance ratio of the unlabeled dataset is defined as $\gamma_u = \frac{\max_c M_c}{\min_c M_c}$, without assuming any specific class distribution. We consider three representative settings:
\begin{itemize}
    \item Consistent: $M_1 \geq M_2 \geq \cdots \geq M_C$ with $\gamma_u = \gamma_l$.
    \item Uniform: $M_1 = M_2 = \cdots = M_C$, i.e., $\gamma_u = 1$.
    \item Reversed: $M_1 \leq M_2 \leq \cdots \leq M_C$, i.e., $\gamma_u = 1/\gamma_l$.
\end{itemize}

To simulate the open-world setting, we introduce the COCO~\citep{lin2014microsoft} dataset as OOD source. COCO contains a diverse set of object categories that are semantically disjoint from those in the target classification task, making it a suitable candidate for evaluating OOD robustness. We mix the COCO dataset with the current unlabeled set to form a more realistic and challenging unlabeled pool, which better reflects the distributional uncertainty encountered in open-world scenarios. We set $t_{HC}=0.95$ for all datasets.

We compare LoFT and LoFT-OW with FixMatch~\cite{sohn2020fixmatch}, as well as equiped with different methods, ACR~\citep{Wei_2023_CVPR}, ACR+BEM~\citep{zheng2024bem}, TCBC~\citep{li2024twice}, CPE~\citep{ma2024three}, and CCL~\citep{zhou2024continuous}. To ensure a comprehensive evaluation, we validate our method on two foundation model backbones: CLIP~\cite{radford2021learning} and OpenCLIP\cite{cherti2023reproducible}, assessing its robustness and generalizability. All experiments are performed on a single NVIDIA A40 GPU. More hyper-parameter settings of our method are in the Appendix.

\subsection{Results on LoFT}

\paragraph{CIFAR-100-LT} 
As shown in Tab.~\ref{cifar_res},  LoFT consistently outperforms PEFT across all settings on CIFAR-100-LT, using both CLIP and OpenCLIP backbones. With OpenCLIP, LoFT achieves the best results in all cases (up to 83.2\%), demonstrating its effectiveness. In terms of imbalance levels, LoFT performs well under all $\gamma$ values. Performance slightly decreases as $\gamma$ increases (e.g., from $\gamma=10$ to $\gamma=20$), indicating increased difficulty with more severe imbalance, but LoFT still maintains a clear margin over PEFT. Moreover, LoFT remains robust under uniform and reversed unlabeled distributions ($\gamma_u=1$ and $1/10$), further validating its ability to handle various class distributions.

\begin{figure}[t]
  \centering
{\includegraphics[width=0.5\textwidth]{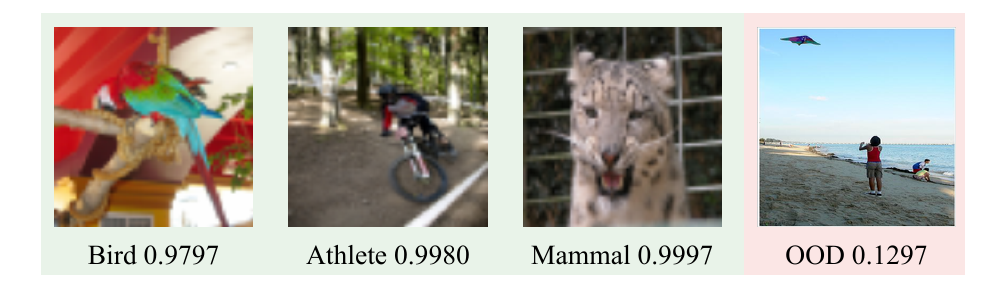}}
    % \vskip -0.05in
   \caption{Visualizations of unlabeled samples and their predicted confidence scores on ImageNet-127. Samples with a green background are assigned reliable pseudo-labels with high confidence, while the sample with a red background is identified as an OOD instance. } 
   \label{example}
   % \vskip -0.05in
\end{figure}

\paragraph{ImageNet-127} 
As shown in Tab.~\ref{imagenet_res}, our method outperforms other methods on a large-scale long-tailed dataset, demonstrating the strong generalization ability of LoFT. Compared to PEFT, LoFT consistently achieves higher accuracy with both CLIP and OpenCLIP backbones, reaching 73.3\% and 73.9\%, respectively. These improvements over strong baselines and prior methods (e.g., FixMatch+CCL at 67.8\%) highlight LoFT's effectiveness beyond small-scale datasets, confirming its robustness and scalability in real-world, large-scale LTSSL scenarios. Moreover, we visualize the unlabeled samples and their prediction scores, as shown in Fig.~\ref{example}. For samples containing meaningful content within the label space, LoFT-OW generates reliable pseudo-labels. In contrast, for uninformative OOD samples, LoFT-OW assigns low confidence scores, facilitating their detection.

\subsection{Results on LoFT-OW}
As shown in Tab.~\ref{cifar_res} and Tab.~\ref{imagenet_res}, LoFT-OW achieves strong performance on both CIFAR-100-LT and ImageNet-127, with fewer training iterations and less data. While its accuracy on CIFAR-100-LT is slightly lower than LoFT due to the inclusion of OOD unlabeled data, which introduces distributional shifts and may hinder representation learning, LoFT-OW remains competitive across all imbalance settings. Notably, on the larger and more complex ImageNet-127 dataset, LoFT-OW outperforms all baselines, including LoFT, demonstrating its superior scalability. This highlights the effectiveness of LoFT-OW in leveraging OOD data when generalizing to more diverse and large-scale benchmarks.

\subsection{Sensitivity Analysis}
We perform two experiments on the CIFAR-100-LT benchmark (N = 50, M = 400, imbalance ratio = 10), using CLIP as our foundation model.

\paragraph{Effect of the hyper-parameter $c_u$}
The hyper-parameter $c_u$ controls the balance between hard and soft pseudo-label assignments. With a large value of $c_u$, more unlabeled samples are assigned hard pseudo-labels, encouraging confident and deterministic supervision but potentially introducing noise if the predictions are incorrect. In contrast, a smaller value of $c_u$ leads to a greater proportion of soft pseudo-labels, which provides more nuanced guidance by preserving model uncertainty, thereby reducing the risk of reinforcing incorrect predictions. As shown in Fig.~\ref{hard_acc}, the test accuracy rises from 74.0\,\% at $c_u=0.2$ to a maximum of 78.8\,\% at $c_u=0.6$, then declines to 75.3\,\% at $c_u=0.95$. This behavior indicates that a moderate confidence cutoff best balances the benefit of incorporating pseudo-labels with the risk of introducing erroneous predictions.

% \begin{figure}[t]
%   \centering
% {\includegraphics[width=0.46\textwidth]{figures/hard_confidence_acc.png}}
%    \caption{Ablation studies on hyper-parameter $c_u$. The horizontal axis represents the value of $c_u$, and the vertical axis represents the accuracy.} 
%    \label{hard_acc}
%    \vskip -0.1in
% \end{figure}

% \begin{figure}[t]
%   \centering
% {\includegraphics[width=0.46\textwidth]{figures/ood_acc.png}}
%    \caption{Ablation studies on hyper-parameter $c_{ood}$. The horizontal axis represents the value of $c_{ood}$, and the vertical axis represents the accuracy.} 
%    \label{ood_acc}
%    \vskip -0.1in
% \end{figure}

\begin{figure}[t]
    \centering
    
    \includegraphics[width=0.48\textwidth]{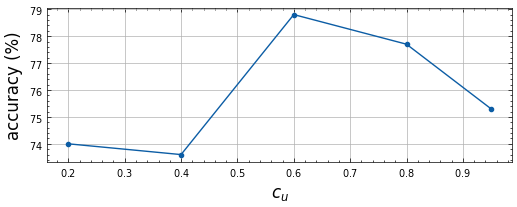}
    
    \vspace{0.1cm} % 调整两张图之间的垂直间距
    
    \includegraphics[width=0.48\textwidth]{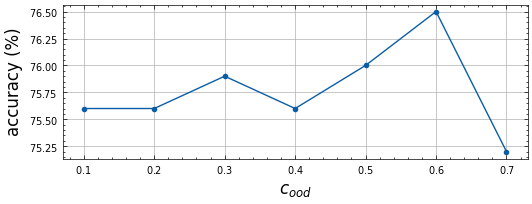}
    
    \caption{Ablation studies on hyper-parameters $c_u$ (top) and $c_{ood}$ (bottom). The horizontal axes represent the values of the respective hyper-parameters, and the vertical axes represent the accuracy.}
    \label{fig:ablation_studies}
    \vskip -0.2in
\end{figure}

\paragraph{Effect of the hyper-parameter $c_{ood}$} The hyper-parameter $c_{ood}$ controls the sensitivity of OOD detection among unlabeled samples. A larger value of $c_{ood}$ enforces stricter filtering, ensuring higher quality among the retained samples but resulting in fewer valid pseudo-labeled instances. Conversely, a smaller $c_{ood}$ allows more samples to pass the filter, increasing quantity but potentially compromising quality due to the inclusion of OOD data. Fig.~\ref{ood_acc} shows that accuracy improves from 75.6\,\% at $c_{\text{ood}}=0.1$ to 76.5\,\% at $c_{\text{ood}}=0.6$ before falling to 75.2\,\% at $c_{\text{ood}}=0.7$. These results suggest that a moderate OOD cutoff effectively excludes OOD samples without discarding too much valuable unlabeled data. Combined with previous experiments, both $c_u$ and $c_{ood}$  present the optimal result at the value of $0.6$. In the standard LTSSL scenario, $c_u=0.6$ corresponds to a confidence level high enough to regard predictions as reliable pseudo-labels. In the open-world setting, $c_{ood}=0.6$ similarly acts as a boundary above which samples are very likely to be in-distribution, thus improving data filtering.

% \section{Conclusion}
% In this work, we revisit LTSSL and propose LoFT, a parameter-efficient framework built on transformer-based foundation models. LoFT tackles key LTSSL challenges such as overconfidence, poor early pseudo-labels, and tail class inefficiency. Leveraging pre-trained models, it enhances calibration, reduces training overhead, and improves pseudo-label quality. We further extend LTSSL to open-world settings with LoFT-OW, which incorporates OOD detection to filter irrelevant samples. Extensive experiments show that LoFT performs competitively in both standard and open-world LTSSL, offering a practical solution for real-world imbalanced learning.

% \section{Conclusion}

% In this work, we address the limitations of training-from-scratch paradigms in LTSSL by proposing \textbf{LoFT}. Theoretically, we find that leveraging FMs allows samller BPE, while their inherent feature compactness strictly compresses the acceptance region for OOD samples. Guided by these insights, we further introduce \textbf{LoFT-OW}, which incorporates a zero-shot filtering mechanism for open-world scenarios. Extensive experiments demonstrate that our framework achieves state-of-the-art performance and superior robustness, effectively utilizing OOD data even with significantly reduced unlabeled samples. Our work solve the problems arising from training from scratch, establishing a new standard for robust imbalanced learning.
\section{Conclusion}

In this work, we propose \textbf{LoFT} to address the limitations of training-from-scratch paradigms in LTSSL. Theoretically, we show that fine-tuning FMs reduces the BPE and enforces feature compactness, which strictly compresses the acceptance region for OOD samples. Guided by these insights, we introduce \textbf{LoFT-OW}, utilizing a dual-stage filtering mechanism for open-world scenarios. Extensive experiments demonstrate that our framework achieves state-of-the-art performance and superior robustness, establishing a new standard for robust imbalanced learning.

\section*{Impact Statements}
This paper presents work whose goal is to advance the field of machine learning. There are many potential societal consequences of our work, none of which we feel must be specifically highlighted here.

\nocite{langley00}

\bibliography{example_paper}
\bibliographystyle{icml2026}

%%%%%%%%%%%%%%%%%%%%%%%%%%%%%%%%%%%%%%%%%%%%%%%%%%%%%%%%%%%%%%%%%%%%%%%%%%%%%%%
%%%%%%%%%%%%%%%%%%%%%%%%%%%%%%%%%%%%%%%%%%%%%%%%%%%%%%%%%%%%%%%%%%%%%%%%%%%%%%%
% APPENDIX
%%%%%%%%%%%%%%%%%%%%%%%%%%%%%%%%%%%%%%%%%%%%%%%%%%%%%%%%%%%%%%%%%%%%%%%%%%%%%%%
%%%%%%%%%%%%%%%%%%%%%%%%%%%%%%%%%%%%%%%%%%%%%%%%%%%%%%%%%%%%%%%%%%%%%%%%%%%%%%%
\newpage
\appendix
\onecolumn

\section{Detailed Theoretical Proofs}
\label{sec:appendix_proofs}

In this section, we provide the detailed mathematical formulations, assumptions, and proofs for the theoretical claims presented in Section~\ref{sec:theory} of the main paper.

\subsection{Generalization Analysis (Lemma~\ref{lemma:gen_bound})}

We analyze the generalization error through the lens of Rademacher Complexity.

\subsubsection{Preliminaries and Definitions}

Let $\mathcal{X}$ be the input space and $\mathcal{Y} = \{1, \dots, K\}$ be the label space. A hypothesis $h: \mathcal{X} \to \mathbb{R}^K$ belongs to a hypothesis class $\mathcal{H}$. We consider the standard supervised learning setting with a loss function $\ell: \mathcal{Y} \times \mathbb{R}^K \to [0, 1]$ (e.g., bounded cross-entropy or 0-1 loss).

\begin{definition}[Rademacher Complexity]
Let $S = \{x_1, \dots, x_N\}$ be a sample of size $N$ drawn i.i.d. from distribution $\mathcal{D}$. The empirical Rademacher complexity of a hypothesis class $\mathcal{H}$ is defined as:
\begin{equation}
    \hat{\mathfrak{R}}_S(\mathcal{H}) = \mathbb{E}_{\sigma} \left[ \sup_{h \in \mathcal{H}} \frac{1}{N} \sum_{i=1}^N \sigma_i \ell(h(x_i), y_i) \right],
\end{equation}
where $\sigma_i$ are independent Rademacher variables taking values $\{-1, +1\}$ with equal probability.
\end{definition}

\begin{theorem}[Generalization Bound via Rademacher Complexity~\citep{bartlett2002rademacher}]
\label{thm:rademacher_general}
For any $\delta > 0$, with probability at least $1-\delta$ over the draw of sample $S$ of size $N$, for all $h \in \mathcal{H}$:
\begin{equation}
    \mathcal{R}(h) \le \hat{\mathcal{R}}_S(h) + 2\hat{\mathfrak{R}}_S(\mathcal{H}) + 3\sqrt{\frac{\ln(2/\delta)}{2N}},
\end{equation}
where $\mathcal{R}(h) = \mathbb{E}[\ell(h(x), y)]$ is the expected risk and $\hat{\mathcal{R}}_S(h)$ is the empirical risk.
\end{theorem}

\subsubsection{Proof of Lemma~\ref{lemma:gen_bound}}

\paragraph{Assumptions on Hypothesis Spaces.}
\begin{itemize}
    \item Let $\mathcal{H}_{scr}$ represent the hypothesis space of training a deep neural network from scratch. This involves optimizing all parameters $W \in \mathbb{R}^D$, where $D$ is very large.
    \item Let $\mathcal{H}_{peft}$ represent the hypothesis space of Fine-tuning a Foundation Model (FM) via PEFT (e.g., LoRA or Adapter). Here, the backbone weights $\theta_{pre}$ are frozen, and only a small set of parameters $\phi \in \mathbb{R}^d$ ($d \ll D$) are optimized.
    \item Consequently, we assume $\mathcal{H}_{peft} \subset \mathcal{H}_{scr}$ (conceptually), or strictly that the effective capacity satisfies $\hat{\mathfrak{R}}_S(\mathcal{H}_{peft}) \ll \hat{\mathfrak{R}}_S(\mathcal{H}_{scr})$.
\end{itemize}

\paragraph{Class-Conditional Bound.}
In Long-Tailed Learning, we analyze the risk for a specific class $y$. Let $S_y$ be the subset of samples belonging to class $y$, with size $N_y = |S_y|$.
Applying Theorem~\ref{thm:rademacher_general} conditionally on class $y$:

\textbf{1. Case: Training from Scratch ($h_{scr} \in \mathcal{H}_{scr}$)}
\begin{equation}
    \mathcal{R}(h_{scr}|y) \le \hat{\mathcal{R}}_{S_y}(h_{scr}|y) + \underbrace{2\hat{\mathfrak{R}}_{S_y}(\mathcal{H}_{scr})}_{\text{Complexity Term}} + \underbrace{3\sqrt{\frac{\ln(2/\delta)}{2N_y}}}_{\text{Sample Size Term}}.
\end{equation}
For tail classes, $N_y \to 0$. Since $\mathcal{H}_{scr}$ is a deep network with massive capacity, $\hat{\mathfrak{R}}_{S_y}(\mathcal{H}_{scr})$ is large (typically scaling with the spectral norms of weight matrices and depth). The ratio $\frac{\mathfrak{R}(\mathcal{H}_{scr})}{\sqrt{N_y}}$ dominates the bound, making it vacuously loose.

\textbf{2. Case: PEFT ($h_{peft} \in \mathcal{H}_{peft}$)}
Since $h_{peft}$ is constrained to a neighborhood of the pre-trained weights, we decompose its risk. Note that the optimal risk achievable in $\mathcal{H}_{peft}$ might be slightly higher than $\mathcal{H}_{scr}$ due to the restricted search space. We define the \textit{Transfer Error} (approximation gap) as $\epsilon_{trans} \approx \min_{h \in \mathcal{H}_{peft}} \mathcal{R}(h) - \min_{h' \in \mathcal{H}_{scr}} \mathcal{R}(h')$.
However, for the specific hypothesis $h_{peft}$ obtained via training, the bound is:
\begin{equation}
    \mathcal{R}(h_{peft}|y) \le \hat{\mathcal{R}}_{S_y}(h_{peft}|y) + 2\hat{\mathfrak{R}}_{S_y}(\mathcal{H}_{peft}) + 3\sqrt{\frac{\ln(2/\delta)}{2N_y}}.
\end{equation}
Crucially, since PEFT only optimizes a few parameters (or low-rank matrices), the complexity term is drastically reduced: $\hat{\mathfrak{R}}_{S_y}(\mathcal{H}_{peft}) \ll \hat{\mathfrak{R}}_{S_y}(\mathcal{H}_{scr})$.
Even if $N_y$ is small, the low numerator keeps the bound tight.

\qed

\subsection{BPE Analysis (Proposition~\ref{thm:loft_main})}

\begin{definition}[Balanced Posterior Error (BPE)]
Following \cite{wei2024learning}, BPE measures the worst-case error across classes. For theoretical analysis, we use the surrogate of worst-case class-conditional risk:
\begin{equation}
    \text{BPE}(h) \approx \max_{y \in \mathcal{Y}} \mathcal{R}(h|y).
\end{equation}
\end{definition}

\paragraph{Proof.}
Let $\mathcal{Y}_{tail}$ be the set of tail classes and $\mathcal{Y}_{head}$ be the set of head classes.
\begin{enumerate}
    \item For $h_{scr}$: In head classes, $N_y$ is large, so the bound is tight and risk is low. In tail classes ($y \in \mathcal{Y}_{tail}$), $N_y$ is small and $\mathfrak{R}(\mathcal{H}_{scr})$ is large. The generalization gap explodes, leading to high expected risk (overfitting). Thus, $\max_y \mathcal{R}(h_{scr}|y)$ is determined by the tail classes.
    
    \item For $h_{peft}$: The complexity $\mathfrak{R}(\mathcal{H}_{peft})$ is small for \textit{all} classes. Provided the Foundation Model features are robust (meaning $\hat{\mathcal{R}}_{S_y}(h_{peft})$ can be minimized during training), the upper bound remains low even for $y \in \mathcal{Y}_{tail}$.
\end{enumerate}

Comparing the worst-case scenarios:
\begin{equation}
    \begin{aligned}
    \text{BPE}(h_{scr}) &= \max_{y} \left( \hat{\mathcal{R}}(h_{scr}|y) + \mathcal{O}\left(\frac{\mathfrak{R}(\mathcal{H}_{scr})}{\sqrt{N_y}}\right) \right) \approx \text{Large (dominated by tail)}, \\
    \text{BPE}(h_{peft}) &= \max_{y} \left( \hat{\mathcal{R}}(h_{peft}|y) + \mathcal{O}\left(\frac{\mathfrak{R}(\mathcal{H}_{peft})}{\sqrt{N_y}}\right) \right) \approx \text{Small}.
    \end{aligned}
\end{equation}
Thus, $\text{BPE}(h_{peft}) < \text{BPE}(h_{scr})$.
\qed

\subsection{OOD Robustness Analysis (Proposition~\ref{prop:ood})}

We model the feature space as a unit hypersphere $\mathbb{S}^{d-1}$, which is a common assumption for normalized embeddings (e.g., Cosine Classifier, CLIP features).

\subsubsection{Geometric Setup}
\begin{itemize}
    \item Let $\mu_k \in \mathbb{S}^{d-1}$ be the prototype (centroid) for class $k$.
    \item A sample $x$ is classified as class $k$ if $\cos(x, \mu_k) \ge t_k$, where $t_k = \cos \theta_k$ is the decision threshold.
    \item The acceptance region is a \textit{Spherical Cap}: $\text{Cap}(\mu_k, \theta_k) = \{x \in \mathbb{S}^{d-1} \mid \langle x, \mu_k \rangle \ge \cos \theta_k\}$.
\end{itemize}

\subsubsection{Concentration of Measure}
We assume OOD samples $x_{out}$ are uniformly distributed on the sphere (maximum entropy assumption for unknown noise). The probability of an OOD sample being falsely accepted by class $k$ is the ratio of the cap area to the sphere area.

\begin{lemma}[Concentration on the Sphere \citep{vershynin2018high}]
For any vector $\mu$ on the unit sphere $\mathbb{S}^{d-1}$ and any $\epsilon \in (0, 1)$:
\begin{equation}
    \mathbb{P}(|\langle x_{out}, \mu \rangle| \ge \epsilon) \le 2 \exp\left(- \frac{d \epsilon^2}{2} \right).
\end{equation}
Considering only the positive direction (the cap), for threshold $t_k = \cos \theta_k$:
\begin{equation}
    \mathbb{P}(x_{out} \in \text{Cap}(\mu_k, \theta_k)) \le \exp\left(- \frac{d \cos^2 \theta_k}{2} \right).
\end{equation}
\end{lemma}

\subsubsection{Comparison: Scratch vs. FM+PEFT}

\textbf{1. Scratch Model ($h_{scr}$):}
Models trained from scratch on long-tailed data suffer from loose decision boundaries for tail classes due to lack of negative sampling constraints. This implies a large angular acceptance $\theta_k^{scr}$ (small $\cos \theta_k^{scr}$).
\begin{equation}
    P_{err}^{scr} \propto \exp\left(- \frac{d}{2} (\cos \theta_k^{scr})^2 \right).
\end{equation}

\textbf{2. FM + PEFT ($h_{peft}$):}
Foundation Models pre-trained with contrastive loss (e.g., InfoNCE) explicitly optimize for alignment (concentration) and uniformity. This results in highly compact intra-class distributions. Fine-tuning with PEFT preserves this geometry. Thus, the angular spread is small: $\theta_k^{peft} \ll \theta_k^{scr}$.
This implies $\cos \theta_k^{peft} \to 1$.

\textbf{Result:}
Since the probability decays exponentially with the square of the cosine threshold, a smaller angle (larger cosine) leads to a massive reduction in error probability.
\begin{equation}
    \frac{\mathbb{P}(x_{out} \in \text{Cap}(\theta_k^{peft}))}{\mathbb{P}(x_{out} \in \text{Cap}(\theta_k^{scr}))} \approx \exp\left( -\frac{d}{2} [\cos^2 \theta_k^{peft} - \cos^2 \theta_k^{scr}] \right) \ll 1.
\end{equation}
This proves that the geometric compactness of PEFT features inherently rejects OOD noise.
\qed

\begin{figure*}[t]
  \centering
{\includegraphics[width=0.8\textwidth]{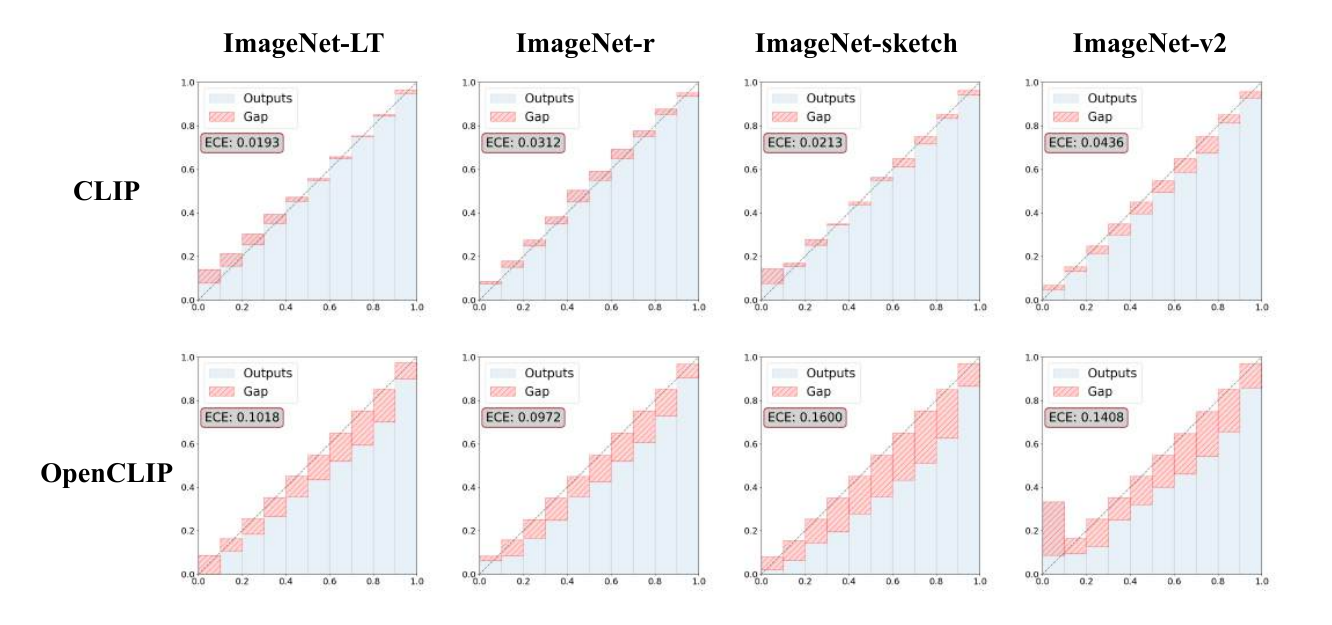}}
   \caption{Zero-shot confidence–accuracy curves across multiple datasets.}
   \label{cofidence}
   % \vskip -0.1in
\end{figure*}

\section{Code}
Detailed code is in the supplementary materials.

\section{Experimental Details}

All experiments are conducted by fine-tuning both CLIP and OpenCLIP as backbone models. We integrate the AdaptFormer modules into every Transformer block. Stochastic Gradient Descent (SGD) is employed as the optimizer, with an initial learning rate of 0.01 scheduled via cosine annealing.

\subsection{CIFAR100-LT}

For the relatively small CIFAR100-LT dataset, we set the total number of optimization steps to 1,024. The learning rate is updated with a cosine annealing schedule every 32 steps.  
\begin{itemize}
  \item \textbf{SSL setting:} $\lambda_{1}=3.0$, $\lambda_{2}=0.0$, and confidence threshold $c_{u}=0.6$.
  \item \textbf{SSL-OW setting:} $\lambda_{1}=2.0$, $\lambda_{2}=1.0$, $c_{u}=0.95$, and out-of-distribution threshold $c_{\mathrm{ood}}=0.6$.
\end{itemize}

\subsection{ImageNet-127}

Given the large scale of ImageNet-127 and the strong representational power of FMs, we sample only 1\% of the training images for our fine-tuning. This still outperforms scratch-trained baselines that use 10\% of the data. We set the total number of optimization steps to 10,000, updating the learning rate via cosine annealing every 100 steps.  
\begin{itemize}
  \item \textbf{SSL setting:} $\lambda_{1}=3.0$, $\lambda_{2}=1.0$, and $c_{u}=0.6$.
  \item \textbf{SSL-OW setting:} $\lambda_{1}=2.0$, $\lambda_{2}=1.0$, $c_{u}=0.95$, and $c_{\mathrm{ood}}=0.6$.
\end{itemize}

\section{Additional Confidence Calibration Results}

To demonstrate the efficiency of our method in the SSL-OW setting, we first perform OOD filtering on the unlabeled dataset, leveraging the foundation model’s reliable confidence estimates. To validate the accuracy of our zero-shot confidence scores, we further visualize confidence–accuracy curves across several datasets.  

Based on the Expected Calibration Error (ECE), we observe that—even in the zero-shot scenario—the foundation model’s confidence estimates remain highly accurate, providing strong justification for using these scores to guide OOD filtering.

% \section{Computational Complexity}
% LoFT is significantly more efficient than traditional methods in terms of training costs

% \begin{itemize}
%     \item Training Time: As shown in Table \ref{imagenet_res}, LoFT converges in only 10,000 iterations (about 4 hours), whereas standard methods (e.g., FixMatch+ACR) require 250,000 iterations (about 8 hours). This yields a 2x speedup.
%     \item Parameters \& FLOPs: Using CLIP-ViT-B/16, LoFT has 149.80M total parameters but only updates 0.18M trainable parameters (vs. WideResNet-28-8 baseline updating all 23.40M). Although the foundation model has higher FLOPs per pass (16.89G vs. 3.37G), the drastic reduction in required training iterations (1/25th of baseline) results in significantly lower overall computational cost for training.
% \end{itemize}

\section{Computational Complexity}

Our proposed method, LoFT, demonstrates significantly superior efficiency compared to traditional semi-supervised learning frameworks in terms of training costs and resource utilization.

\vspace{0.5em}
\noindent\textbf{Training Efficiency and Convergence.} 
As summarized in Table \ref{imagenet_res}, LoFT exhibits rapid convergence properties. Specifically, our model reaches optimal performance within only 10,000 iterations (requiring approximately 4 hours). in stark contrast, standard methods such as FixMatch combined with ACR require up to 250,000 iterations (approximately 8 hours) to achieve comparable results. Consequently, LoFT delivers a $\mathbf{2\times}$ speedup in total training time while reducing the number of required update steps by a factor of 25.

\vspace{0.5em}
\noindent\textbf{Parameter Efficiency and FLOPs.} 
We further analyze the computational overhead in terms of model parameters and Floating Point Operations (FLOPs). Utilizing the CLIP-ViT-B/16 as the foundation model, LoFT contains a total of 149.80M parameters; however, it introduces a highly efficient fine-tuning strategy that updates only \textbf{0.18M} trainable parameters. This is significantly more efficient than the WideResNet-28-8 baseline, which requires updating all 23.40M parameters. Although the large-scale foundation model incurs higher computational costs per forward pass (16.89G FLOPs vs. 3.37G FLOPs for WideResNet), the drastic reduction in total training iterations results in a significantly lower aggregate computational cost. This makes LoFT not only faster but also more computationally economical for practical deployment.

%%%%%%%%%%%%%%%%%%%%%%%%%%%%%%%%%%%%%%%%%%%%%%%%%%%%%%%%%%%%%%%%%%%%%%%%%%%%%%%
%%%%%%%%%%%%%%%%%%%%%%%%%%%%%%%%%%%%%%%%%%%%%%%%%%%%%%%%%%%%%%%%%%%%%%%%%%%%%%%

\end{document}